\documentclass[10pt, a4paper]{article}

\usepackage{lrec-coling2024} 

\title{DP-CRE:  Continual Relation Extraction via Decoupled Contrastive Learning and Memory Structure Preservation}
         
\name{Mengyi Huang$^{1,2,\dagger}$, Meng Xiao$^{1,\dagger}$, Ludi Wang$^1$, Yi Du$^{1,2,3,*}$\thanks{$^*$ Corresponding Author}\thanks{$^\dagger$ Equal Contribution}} 

\address{$^1$Computer Network Information Center, Chinese Academy of Sciences
\\$^2$University of Chinese Academy of Sciences
\\$^3$Hangzhou Institute for Advanced Study, UCAS\\
         \{myhaung, shaow, wld, duyi\}@cnic.cn}

\usepackage{graphicx}
\usepackage{tabularx}
\usepackage{soul}
\usepackage{subfig}
\usepackage{amsmath}
\usepackage{upgreek}
\usepackage{algorithm}
\usepackage{algorithmicx}
\usepackage{algpseudocode}

\abstract{
Continuous Relation Extraction (CRE) aims to incrementally learn relation knowledge from a non-stationary stream of data.
Since the introduction of new relational tasks can overshadow previously learned information, catastrophic forgetting becomes a significant challenge in this domain. 
Current replay-based training paradigms prioritize all data uniformly and train memory samples through multiple rounds, which would result in overfitting old tasks and pronounced bias towards new tasks because of the imbalances of the replay set. 
To handle the problem, we introduce the DecouPled CRE (DP-CRE) framework that decouples the process of prior information preservation and new knowledge acquisition. 
This framework examines alterations in the embedding space as new relation classes emerge, distinctly managing the preservation and acquisition of knowledge. 
Extensive experiments show that DP-CRE significantly outperforms other CRE baselines across two datasets.
The code and data are publicly accessible via~\href{https://github.com/mengyi99/DP-CRE}{https://github.com/mengyi99/DP-CRE}.
 \\ \newline  
}

\begin{document}

\maketitleabstract

\section{Introduction}
Relation extraction seeks to discern patterns of relationships between entities within textual data~\cite{zhou2020survey}. 
A significant challenge in deploying this technique arises when new documents continuously emerge, introducing both novel entity types and relation categories. 
A traditional approach involves retraining the model from scratch whenever new data or relations appear, but persistently storing and retraining on every new sample becomes impractical due to constraints in storage and computational resources. 
An alternative method is to incrementally train the model using these new samples. 
Yet, this approach can lead the model to experience catastrophic forgetting and struggle with potential newly introduced relation classes.
Additionally, the domain shift between successive batches of training data can result in a pronounced bias towards recent tasks.


\begin{figure}
\begin{center}
\centering
\includegraphics[width = 0.45\textwidth]{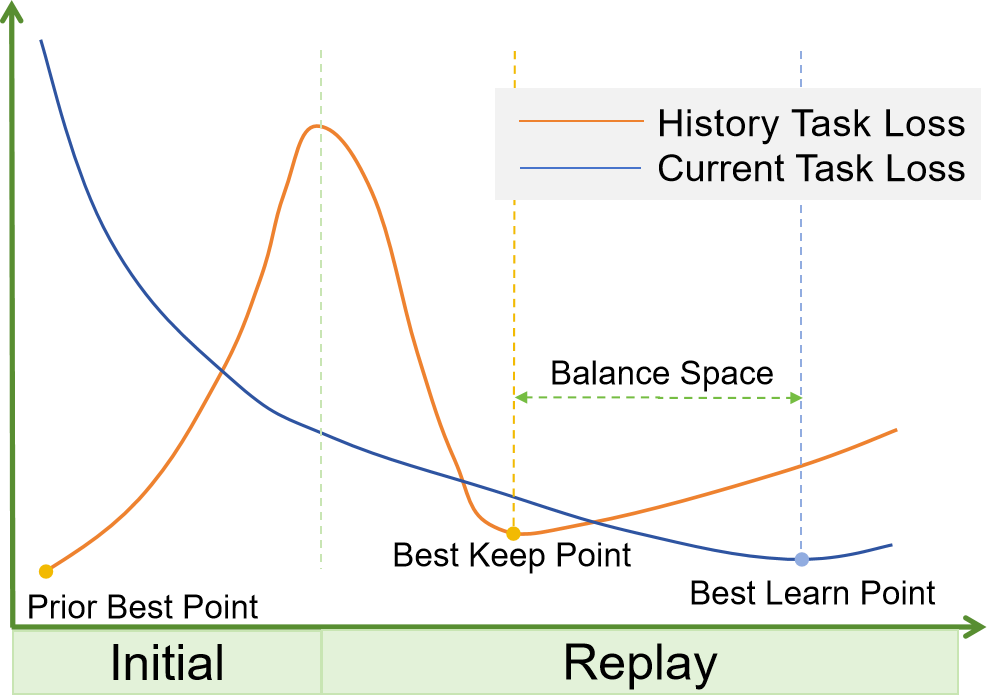} 
\caption{The balance essence of continual relation extraction. Replaying is the period that model parameters compete between learning new data and preserving prior task knowledge.} 
\vspace{-0.6cm}
\label{fig. Balance}
\end{center}
\end{figure} 

\smallskip
Continual relation extraction (CRE)~\cite{wang2019sentence} has been developed to address these challenges, which can be viewed as two tasks: Task \uppercase\expandafter{\romannumeral1}) \textit{Prior Information Preservation} and Task \uppercase\expandafter{\romannumeral2}) \textit{New Knowledge Acquisition}. 
One prevailing approach is memory-based continual learning, tailored to the intricacies of Neural Language Processing (NLP) tasks that necessitate only modest storage for data samples, which is designed to counteract biases by training the model on a combination of prior relation memory samples and new relation samples. 

These memory samples, though limited in number, are meticulously chosen to encapsulate the essence of the original training set. To enhance the efficacy of these memory samples, given their smaller volume compared to the original training set, researchers have proposed additional training strategies on them~\cite{zhao2022consistent,zhao2023improving}. This amplifies their impact on the learning process. Furthermore, legacy information from prior tasks isn't discarded but preserved in older model versions, be it through the model's sample embedding~\cite{cui2021refining} or its parameters~\cite{xia2023enhancing}.
 
\smallskip
The knowledge from previous tasks is encapsulated within the existing model and memory samples, while the insights from new tasks emerge from their respective training. 
Historically, research has intertwined these two tasks during the memory replay learning phase. 
However, as the training progresses through multiple rounds, the representativeness of these memory samples diminishes due to the looming threat of overfitting. 
In each CRE iteration, they're treated equivalently to new samples. This equal treatment can be problematic, as memory samples, having been extensively trained already, might suffer from information dilution. 
In addition, the intricacies of one task can inadvertently influence the other. 
As illustrated in figure\ref{fig. Balance}, before the initial learning of a new task, historical tasks remain in an optimal state since they remain unaffected by new data types.
Yet, during the replay phase, even if the detrimental effects of the new task's initial learning on historical tasks diminish, the theoretical optimum for the training set doesn't necessarily translate to the test set, given the imbalances in the replay set.

\smallskip
To address the aforementioned issues, we introduce the \textbf{D}ecou\textbf{P}led \textbf{C}ontinual \textbf{R}elation \textbf{E}xtraction framework (DP-CRE). 
This framework emphasizes treating memory samples and new samples as separate entities. 
We aim to cluster similar new task samples within the feature space, ensuring clear differentiation between relation labels via decoupled contrastive learning. 
Concurrently, we aspire to preserve the structure between memory samples and keep them distributed evenly to maintain representativeness throughout the training trajectory.

\smallskip
In summary, our contributions can be listed as follows:

\noindent (1) \textbf{Balancing CRE with Multi-task Learning}: By categorizing CRE into Prior Information Preservation and New Knowledge Acquisition, DP-CRE can facilitate a more targeted approach to each task, eliminating the complexities that arise from their conflation. We explore the multi-task learning task and update the model to achieve better performance for both tasks simultaneously. 

\noindent (2) \textbf{Decoupling to Mitigate Overfitting}: To address the overfitting issue stemming from repetitive training on memory samples, we adopt a decoupled processing approach for old and new samples. We also introduce a method to conserve the memory structural information by restricting the change amount of embedding to ensure representativeness.

\noindent (3) \textbf{Empirical Validation of DP-CRE}: We conduct extensive experiments to show the superiority of the proposed method. The experimental results demonstrate that our model achieves state-of-the-art accuracy compared with existing works.

\section{Related Work}

\smallskip
\noindent\textbf{Continual Learning} can be divided into three main categories. (1) Regularization-based method~\cite{li2017learning,kirkpatrick2017overcoming, YangSZFZXY23} introduces regularization terms in training loss to avoid overfitting and excessive adjustment of the model parameters. (2) Architecture-based method~\cite{fernando2017pathnet,mallya2018packnet,YangZZX019} adapts the model architecture dynamically to learn new tasks without forgetting previous tasks. (3) Memory-based method~\cite{rebuffi2017icarl,lopez2017gradient} stores representative old task data and preserves old task knowledge by replaying stored samples or generating data through generative methods. 

\smallskip
The memory-based method is widely used in current \textbf{Continual Relation Extraction} work and shows better performance than the other two categories. The quintessential memory-based CRE methodology, as outlined in~\cite{han2020continual}, segments CRE into four distinct phases:
\uppercase\expandafter{\romannumeral1}) \textit{Initial Learning};
\uppercase\expandafter{\romannumeral2}) \textit{Selection of Representative Samples};
\uppercase\expandafter{\romannumeral3}) \textit{Memory Replay Learning}; and
\uppercase\expandafter{\romannumeral4}) \textit{Joint Prediction}.
Besides, numerous studies have sought to refine and enhance this foundational approach. 
For instance, research focused on phase \uppercase\expandafter{\romannumeral1} emphasizes the comprehensive acquisition of new task knowledge~\cite{wang2022learning,xia2023enhancing}, while investigations centered on phase \uppercase\expandafter{\romannumeral3} aim to optimize the memory replay process to mitigate forgetting~\cite{wu2021curriculum,cui2021refining,hu2022improving,zhao2022consistent,zhang2022prompt,zhao2023improving}.
However, these methods train memory samples and new task samples with the same status, which would bring model bias. ~\cite{wang2022less} attempts to balance tasks simply by reducing the number of new samples,  which causes the disadvantage of losing the opportunity to compare old samples with most new samples.

\smallskip
\noindent\textbf{Contrastive Learning} is a method of self-supervised learning to increase the distinguishability of different classes of samples in the feature space. In the CRE task, ~\cite{zhao2022consistent} uses contrastive learning on the distribution of prototypes. ~\cite{hu2022improving} adds a contrastive network to guide the embedding at the memory replaying stage by rewarding the closeness of prototypes and their positive memory samples. However, the excessive replay-learning process may cause a disuniform distribution of memory samples, so that the calculated prototypes could not accurately represent the relation. ~\cite{zhao2023improving} utilizes the previous model and limits the memory samples to the same location in the feature space to ensure new relations do not impact how prior relations are embedded. The approach can enhance the consistency of the model's performance, but may also limit the ability to learn new relations. When the model is restricted to a particular set of positions in the memory space, it may not be able to generalize its learning to new patterns. 

\section{Task Formulation}
In continual relation extraction, there are successive tasks $(T^1, T^2, ..., T^k)$ with each task $T^i$ containing triplets as $(R^i, D^i, Q^i)$. Here, $R^i$ represents the set of new relations, and $D^i$ and $Q^i$ represent the training and testing sets, respectively. An instance $(x_i, y_i)$ in $D^i \cup Q^i$ is a sentence $x_i$ and its corresponding relation $y_i$. The first occurrence of the training data $D^i$ containing new relations happens only during the training phase of task $T^i$. During the testing phase of task $T^i$, all previous testing sets $(Q^1, Q^2, ..., Q^i)$ are required.
In the subsequent training process, samples in the memory will be replayed to alleviate catastrophic forgetting. After training, only limited data is saved in $M = M^1\cup M^2\cup...\cup M^i$ due to memory limitation.

\begin{figure*}
\begin{center}
\includegraphics[width = 0.85\textwidth]{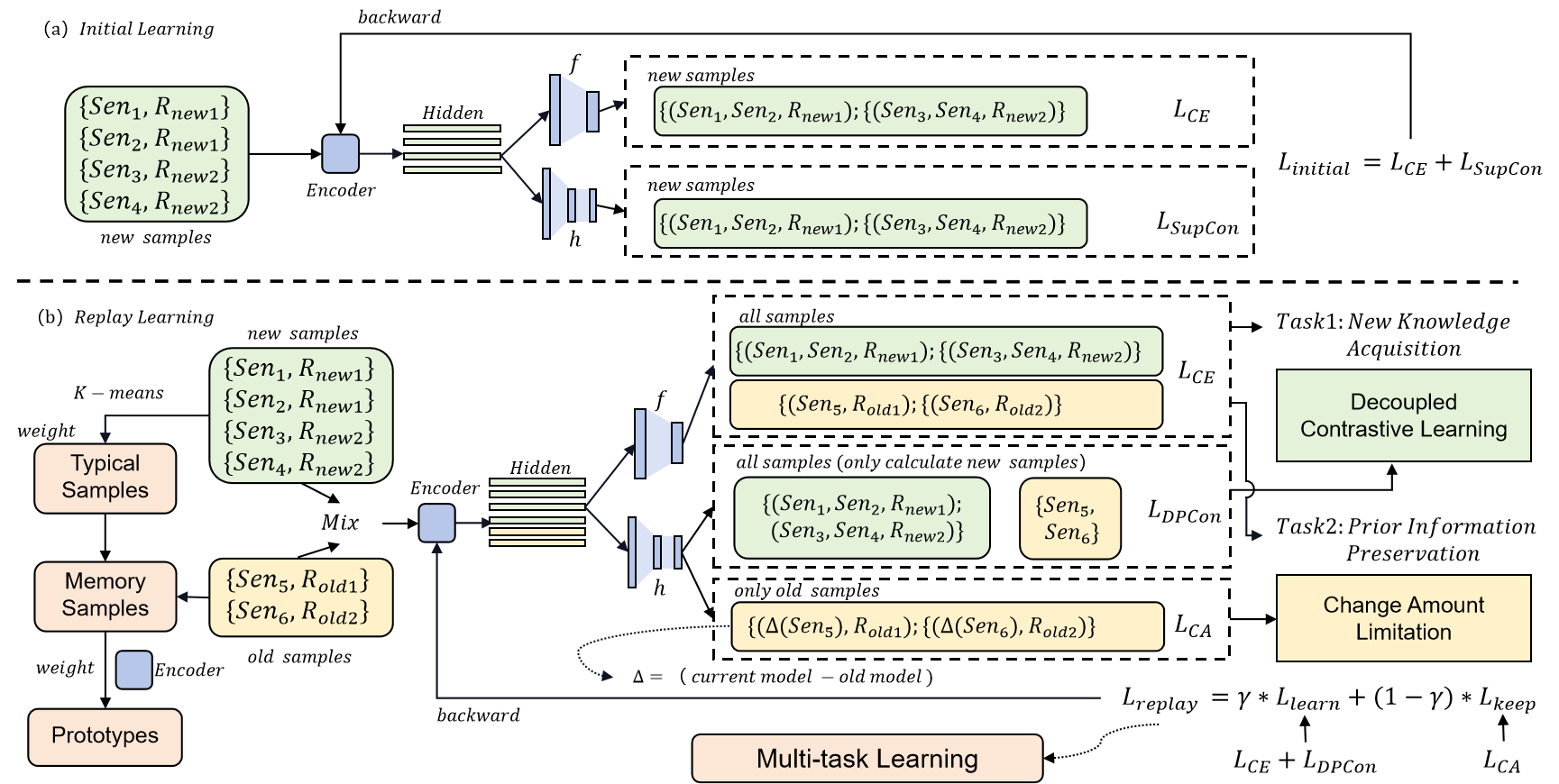} 
\caption{DecouPled Framework of DP-CRE for $T_k$. Green cubes represent prior tasks and yellow cubes represent new tasks. (a) Initial Learning is the routine training on new samples. (b) Replay Learning balances New Knowledge Acquisition and Prior Information Preservation using \textbf{DecouPled Contrastive Learning} and \textbf{Change Amount Limitation}.} 
\label{fig. Model overview}
\end{center}
\vspace{-0.6cm}
\end{figure*}

\section{DecouPled Framework}
\subsection{Model Design}
The model consists of a shared embedding layer and two separate classifier layers. The shared embedding layer $E$ includes BERT~\cite{kenton2019bert} embedding network and a simple FNN network to encode sentences into feature space. 
For a sentence $x_i$ in $D^k$ with relation label $y_i = r \in R^k$ of $T^k$, the embedding layer encoder $x_i$ into a high-dimensional vector $z_i$.
\begin{equation}
\begin{aligned}
    z_i =& E(x_i)\\
\end{aligned}
\end{equation}
Classifier layers $C$ include a classification head and a contrastive head. Through the classification head, $z_i$ is embedded to $f(z_i)$:
\begin{equation}
\begin{aligned}
    f(z_i) = W_1(z_i)+b_1, \\
\end{aligned}
\end{equation}
where $W_1$ and $b_1$ are trainable parameters to extract sample classification features with cross-entropy loss $L_{ce}$:
\begin{equation}
\begin{aligned}
    L_{ce} =& \sum_{i\in D^k} \frac{-1}{|D^k|} \sum_{r\in R^k}\delta_{y_i=l_r} \\
    &\times log {\frac{exp(f(z_i),\boldsymbol{l_r}))}{\sum_{r\in R^k}exp(f(z_i),\boldsymbol{l_r}))}},\\
\end{aligned}
\end{equation}
where $\delta_{y_i=l_j}=1$ when $l_r$ is real relation label of sentence $x_i$, otherwise $\delta_{y_i=l_j}=0$.

Through the contrastive head ,$z_i$ is embedded to $h(z_i)$:
\begin{equation}
\begin{aligned}
    h(z_i) = W_3(ReLU(W_2(z_i)+b_2))+b_3, \\
\end{aligned}
\end{equation}
where $W_2, W_3$ and $b_2,b_3$ are trainable parameters for dimension reduction. We train the model with Supervised Contrastive Loss~\cite{khosla2020supervised}. For each anchor sample, randomly select one positive sample within the same category and negative samples from different categories in the same batch to calculate $L_{SupCon}$:
\begin{equation}
\begin{aligned}
     L_{SupCon} =& \sum_{i\in D^k} \frac{-1}{|D^k|} \sum_{j\in D^k}\delta_{y_i=y_j} \\
    &\times log {\frac{exp\left((h(z_i) \cdot h(z_j)/\uptau\right)}{\sum_{j\in D_k}exp\left((h(z_i) \cdot h(z_j)/\uptau\right)}}, \\
\end{aligned}
\end{equation}
where $\uptau$ is the temperature coefficient.

\renewcommand{\algorithmicrequire}{\textbf{Input:}} 
\renewcommand{\algorithmicensure}{\textbf{Output:}} 
\begin{algorithm}
    \caption{Train DP-CRE for the $T^k$} 
    \begin{algorithmic}[1] 
        \Require $E^{k-1}$,$C^{k-1}$,$(R^k,D^k)$,$M$,
        \Ensure  $E^{k}$ , $C^{k}$,$M$ ,$P_r$
        \State $E^{k}\Leftarrow E^{k-1}, C^{k}\Leftarrow C^{k-1}$
        \For{$i = 1 \to epoch_1$} 
            \State Calculate $L_{ce},L_{SupCon}$ with $D_k$
            \State $L_{initial}\Leftarrow L_{ce}+L_{SupCon}$
            \State Update $E^{k},C^{k}$ with $\nabla L_{initial}$  
        \EndFor
        \For{$i = 1 \to epoch_2$} 
            \State Calculate $L_{ce},L_{DPCon}$ with $D_k,M$
            \State $L_{learn}\Leftarrow L_{ce}+L_{DPCon}$   
            \State Calculate $L_{CA}$ with $M,E^{k-1},C^{k-1}$
            \State $L_{keep}\Leftarrow k^{\lambda}\times L_{CA}$ 
            \State Calculate $\boldsymbol{\theta_l},\boldsymbol{\theta_k}$ of $E^{k},C^{k}$ with $\nabla L_{learn},\nabla L_{keep}$
            \If {$\theta_l^T\theta_k\geq\theta_l^T\theta_l$}
            \State $\gamma \Leftarrow 1$
            \ElsIf{$\theta_l^T\theta_k\geq\theta_k^T\theta_k$}
            \State $\gamma \Leftarrow 0$
            \Else
            \State $\gamma \Leftarrow \frac{(\boldsymbol{\theta_k}-\boldsymbol{\theta_l})^T\boldsymbol{\theta_k}}{\Vert\boldsymbol{\theta_l}-\boldsymbol{\theta_k}\Vert_2^2}$
            \EndIf
            \State  $L_{replay} \Leftarrow  \gamma \times L_{learn} + (1-\gamma) \times L_{keep}$
            \State Update $E^{k},C^{k}$ with $\nabla L_{replay}$
        \EndFor
        \For{$r \in |R_k|$} 
            \State Select $M^{r}$ and get $C_{r}$ with K-means on $D_{k}$
            \State $w_{r,i} \Leftarrow \frac{|C_{r,i}|}{\sum_{i}^{|M^r|}|C_{r,i}|}$
            \State $M \Leftarrow M\cup M^{r}$
            \EndFor
            \For{$r \in |M|$} 
            \State $P_r = \sum_{i\in |M_r|}w_{r,i}\cdot z_{i,r})$
            \EndFor
            \State \Return $Encoder^{k},Classifier^{k},M,P_r$ 
    \end{algorithmic}
\end{algorithm}
\vspace{-0.3cm}

\subsection{Initial Learning}
DP-CRE replicates the prior model $E^{k-1}$ and $C^{k-1}$ before new task $T^k$ arrives to control the direction of model training.
 
When new task $T_k$ arrived, we fine-tune the model using new task data $D^k = \left\{\left(x_1^{D^k},y_1^{D^k}\right),...,\left(x_N^{D^k},y_N^{D^k}\right)\right\}$ with the sole purpose of new knowledge acquisition, as illustrated in figure~\ref{fig. Model overview}(a). If we focus too much on retaining prior task information at the beginning, the model's capacity to learn new relations $R^k$ would be hindered. Cross-entropy loss $L_{ce}$ and contrastive learning loss $L_{con}$ are employed concurrently to decrease the distance among similar relation samples in the embedding space.

\begin{equation}
\begin{aligned}
    L_{initial} &= L_{ce} + L_{SupCon}
\end{aligned}
\end{equation}
A certain amount of initial learning is necessary and can improve the model's overall accuracy because the model has already reached the optimal parameters $w^{k-1}$. Initial learning prompts the model to jump out of $w^{k-1}$ and search the optimal parameters of the joint task in a larger space, rather than falling into the local optimum point of previous tasks.

\subsection{Replay Learning}
As shown in~\ref{fig. Model overview}(b), all prior memory samples $M$ and new relations training sets $D^k$ are mixed in the replay learning process. At this stage, DP-CRE regards CRE as the combination of New Knowledge Acquisition and Prior Information Preservation. We minimize the distance between $D^k$ through decoupled contrastive learning and maintain the consistency of $M$ by restricting the embedding change amount to accomplish the purpose of memory structure preservation.

\subsubsection{New Knowledge Acquisition: Decoupled Contrastive Learning}
Similar to initial learning, the first task for replay learning is to acquire new knowledge from $D^k$. We still use the separate classifier layers model and entropy loss $L_{ce}$, but only new task samples to calculate $L_{SupCon}$, which is decoupled contrastive learning of DP-CRE. Memory samples are selected to represent all prior samples, and the embedding of unselected samples is positioned between them. If $L_{SupCon}$ is still applied to memory samples, excessive training can lead to information loss and overfitting. The decoupled $L_{DPCon}$ would not reward the reduction of distance between memory samples: 

\begin{equation}
\begin{aligned}
    L_{ce} =& \sum_{i\in D^k \cup M} \frac{-1}{|D^k \cup M|} \sum_{r\in{|R^k \cup R|}}\delta_{y_i=l_r} \\
    &\times log {\frac{exp(f(z_{i,r}),\boldsymbol{l_r}))}{\sum_{r=1}^{|R^k \cup R|}exp(f(z_{i,r}),\boldsymbol{l_r}))}},\\
     L_{DPCon} =& \sum_{i\in D^k} \frac{-1}{|D^k|} \sum_{j\in D^k}\delta_{y_i=y_j} \\
    &\times log {\frac{exp\left((h(z_{i,r}) \cdot h(z_{j,r})/\uptau\right)}{\sum_{j\in D_k \cup M}exp\left((h(z_{i,r}) \cdot h(z_{j,r})/\uptau \right)}}, \\
    L_{learn} =& L_{ce} + L_{DPCon},
\end{aligned}
\end{equation}
where $\uptau$ is the temperature coefficient and only new relation samples were calculated for in numerator of contrastive loss.

After splitting the new task sample separately, decoupled contrastive learning reduces the distance between new task samples, and memory samples only serve as negative anchors to obtain more accurate and reliable outcomes. As a result, this part is a separate new knowledge acquisition task. 
\begin{figure}
\centering
\subfloat[Untrained Samples in Feature Space]{ \includegraphics[width = 0.23\textwidth]{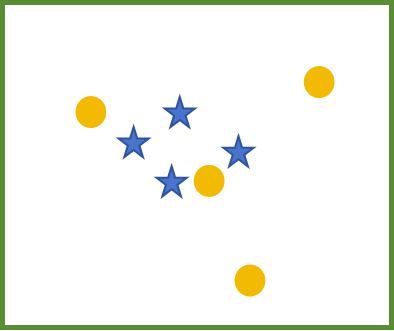}}
\hfill
\subfloat[Contrastive Learning to All Data]
{\includegraphics[width = 0.23\textwidth]{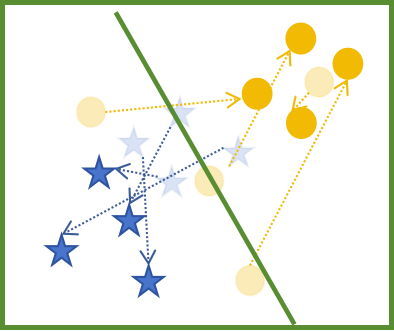}}
\hfill
\subfloat[Retaining Old Samples Unchanged]
{\includegraphics[width = 0.23\textwidth]{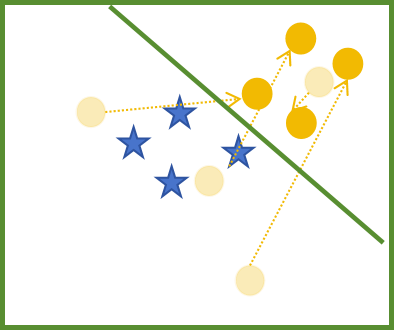}} 
\hfill
\subfloat[Decoupling Old and New Samples]
{\includegraphics[width = 0.23\textwidth]{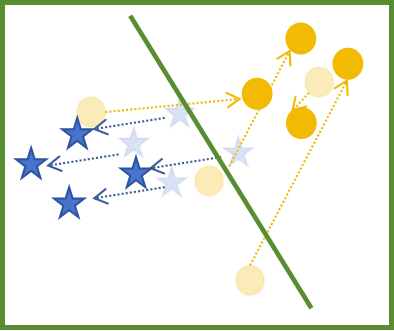}}
\caption{(a) In the feature space, blue pentagrams indicate old samples while yellow circles represent new ones. (b) Applying contrastive learning to all data would destroy the memory structure information. (c)Retaining old samples unchanged would limit the classification ability of the model. (d) Our approach is to decouple old and new samples so that the structure information is preserved by obtaining a better classification boundary.}
\vspace{-0.6cm}
\label{fig. change amount}
\end{figure}
\subsubsection{Prior Information Preservation: Change Amount Limitation}
To ensure the model's ability to maintain prior relations, DP-CRE proposes another separate prior information preservation task that restricts the embedding of old samples. When replay training the model, we use the saved model $E^{k-1}$ and $C^{k-1}$ to guide the process. Previous approaches have controlled memory samples by maintaining the same embedding, but this may restrict the ability to learn new information, as shown in figure~\ref{fig. change amount}. Additionally, it is important to consider the memory structure information between chosen memory samples, which are uniformly distributed in prior well-trained model to ensure their representativeness.

During the replay learning process, DP-CRE puts a limit on the amount of change in similar memory samples between the preserved and the current models. As memory samples with the same label are usually close in the prior model, the change amount limitation, denoted as $L_{CA}$, ensures that related samples remain close in the new model with the same distance, preserving the structure information between them to maintain consistency:

\begin{equation}
\begin{aligned}
    L_{CA} =& \sum_{i,j\in M} \frac{1}{|M|} \delta_{y_i=y_j} \\
    &\times \Vert\left((h^k(z_{i,r}) - h^{k-1}(z^{k-1}_{i,r})\right) - \\
    &\left((h^k(z_{j,r}) - h^{k-1}(z^{k-1}_{j,r})\right)\Vert_2, \\
\end{aligned}
\end{equation}
where $z^{k-1},h^{k-1}$ is previous embedding layer and classifier layer of $E^{k-1}$ and $C^{k-1}$.

Change Amount Limitation is a task that involves memory samples for prior information preservation. In this way, DP-CRE decouples CRE into two separate parts: new and old relations, and allows us to control the proportion between them. We employ an extra module that takes inspiration from multi-task learning to learn the balance point between these two parts.

\subsubsection{Multi-task Balance}

The replay loss is split into two components: $L_{learn}$ to learn new tasks from $D^k$, and $L_{CA}$ to retain learned tasks from $E^{k-1}$, $C^{k-1}$ and $M$. The balance of the model in both the old and new tasks is determined by the inclination toward these two losses. DP-CRE treats it as a multi-task learning work. Following~\cite{sener2018multi}, we calculate the balance parameter $\gamma$ to reach a \textbf{Pareto Optimality}. Additionally, the balance ratio of two tasks is related to the learned relation number since the percentage of keeping prior knowledge in the model grows as the number of learned relations increases:
\begin{equation}
\begin{aligned}
    L_{keep} &= k^{\lambda} \times L_{CA}\\
    \gamma &= \left \{
    \begin{array}{ll}
    1,   & \theta_l^T\theta_k\geq\theta_l^T\theta_l\\
    0,  &\theta_l^T\theta_k\geq\theta_k^T\theta_k\\
    \frac{(\theta_k-\theta_l)^T\theta_k}{\Vert\theta_l-\theta_k\Vert_2^2},  & otherwise
    \end{array}
    \right. \\
    L_{replay} &=  \gamma \times L_{learn} + (1-\gamma) \times L_{keep},\\
\end{aligned}
\end{equation}
where $k$ is the task round in the the experimental setup, $\lambda$ is a hyper-parameter, and $\theta_l$,$\theta_k$ is the gradients of loss $L_{learn}$ and $L_{keep}$.

\subsection{Weighted Prototype and Double-NCM Prediction}
We employ the K-Means algorithm to cluster the embedding of training data for each relation. Then, typical samples $M_r$ which are closest to the cluster centers for $r \in R^k$ are selected with the cluster number depending on the available memory space.

To avoid catastrophic forgetting, these selected samples are retained as memory samples $M_r \to M$ and replayed during the training of new tasks. We perform typical sample selection after the memory replay learning to make full use of all data in each training period. To calculate the prototypes more accurately, the proportion of each cluster to all samples is recorded as the weight of the memorized sample $w_{r, i}$:

\begin{equation}
\begin{aligned}
   w_{r,i} &= \frac{|C_{r,i}|}{\sum_{i}^{|M_r|}|C_{r,i}|}, \\
\end{aligned}
\end{equation}

where $|M_r|$ is the memory samples number of relation $r$ in $R^k$, and $|C_{r,i}|$ is the amount of cluster. 

For each current task relation label $r$, we calculate memory prototype $P_r$ after selecting representative memory samples. 

\begin{equation}
\begin{aligned}
    P_r &= \sum_{i\in M_r}w_{r,i}\cdot z_{i,r} \\
\end{aligned}
\end{equation}

where $z_{i,r}$ is the embedding of sample $x_i$ in memory set $M_r$ with relation label $y_i=r \in R_k$. 

To represent relations, we use the weighted average embedding of memorized samples as the relation prototypes. When presented with a test sample $x$, the nearest class mean (NCM)~\cite{mai2021supervised} algorithm calculates the distances between the embedding of $x$ and all prototypes and then predicts $x$ to the label of the nearest prototype. Additionally, we improve the prediction accuracy by utilizing memory samples.

\begin{equation}
\begin{aligned}
    y^{*}&=\arg\min\limits_{r=1,...,k}\left(\Vert z_{i}-P_r + \min\limits_{j\in M_r}(\Vert z_{i}-z_{j,r}\Vert)\Vert\right),
\end{aligned}
\end{equation}
where $z_i$, $z_{i,r}$ is the embedding of sample $x_i$ in testing set $T_k$ or memory set $M$ with $y_i=r$ and $y^{*}$ is the predicted label. 

\begin{table*}
\centering
\scalebox{0.8}{\begin{tabular}{l|cccccccccc}
\hline
\multicolumn{11}{c}{\textbf{FewRel}}\\
\hline 
\multicolumn{1}{c}{Model}    &$T_1$     &$T_2$    &$T_3$    &$T_4$      &$T_5$    &$T_6$    &$T_7$     &$T_8$    &$T_9$    &$T_{10}$\\
\hline
EA-EMR~\cite{wang2019sentence}&89.0      &69.0     &59.1     &54.2       &47.8     &46.1     &43.1      &40.7     &38.6     &35.2\\
EMAR(BERT)~\cite{han2020continual}&98.2  &94.8     &92.6     &91.1      &89.7     &87.9     &87.1      &86.0     &84.7     &83.3\\
CML~\cite{wu2021curriculum}   &91.2      &74.8     &68.2     &58.2      &53.7     &50.4     &47.8      &44.4     &43.1     &39.7\\
RP-CRE~\cite{cui2021refining} &98.1      &94.8     &92.6     &91.1      &89.7     &87.9     &87.1      &86.0     &84.7     &83.3\\
CR-ECL~\cite{hu2022improving} &97.8 &94.9 &92.7 &90.9 &89.4 &87.5 
&85.7 &84.6 &83.6 &82.7\\
ACA~\cite{wang2022learning}   &\underline{98.4}  &95.1     &93.0     &91.5  &90.5     &88.9     &87.9      &86.7     &\underline{85.8}     &\underline{84.4}\\
CRL~\cite{zhao2022consistent} &98.0  &94.3   &92.4  &90.5  &89.5  &87.8   &87.0   &85.6  &84.3  &83.0 \\
CEAR~\cite{zhao2023improving} &98.3 &\textbf{95.6} &\underline{93.5}     &\underline{92.0}  &\underline{90.8}     &\underline{89.3}     &\underline{88.0} &\underline{86.8} &85.6&84.0\\

\hline
\textbf{Ours}   &\textbf{98.5}  &\underline{95.4}   &\textbf{93.7}     &\textbf{92.1}  &\textbf{90.9}   &\textbf{89.4}   &\textbf{88.5}      &\textbf{87.4}   &\textbf{86.3}  &\textbf{85.1}\\

\hline
\hline
\multicolumn{11}{c}{\textbf{TACRED}}\\
\hline
\multicolumn{1}{c}{Model}    &$T_1$     &$T_2$    &$T_3$    &$T_4$   &$T_5$    &$T_6$    &$T_7$     &$T_8$    &$T_9$    &$T_{10}$\\
\hline
EA-EMR~\cite{wang2019sentence}&47.5      &40.1     &38.3     &29.9     &24.0     &27.3     &26.9      &25.8     &22.9     &19.8\\
EMAR(BERT)~\cite{han2020continual}&98.0  &93.0     &89.7     &84.7    &82.7     &81.5     &79.0      &77.5     &77.6     &77.1\\
CML~\cite{wu2021curriculum}   &57.2      &51.4     &41.3     &39.3      &35.9     &28.9     &27.3      &26.9     &24.8     &23.4\\
RP-CRE~\cite{cui2021refining} &96.6      &91.4     &88.8     &84.8      &82.8     &81.0     &77.9      &77.4     &76.5     &75.7\\
CR-ECL~\cite{hu2022improving} &97.3      &92.5     &88.2     &85.6        &83.7     &83.3     &81.8      &80.1     &77.7     &76.8\\
ACA~\cite{wang2022learning}   &\textbf{98.2}  &\underline{93.8}  &89.9     &85.9      &84.2     &82.7     &80.5      &78.4     &78.6     &77.5\\
CRL~\cite{zhao2022consistent} &\underline{98.0}&\textbf{93.9}  &\underline{90.8}    &86.0 &84.9     &82.9     &80.1      &79.2     &79.4     &78.5\\
CEAR~\cite{zhao2023improving} &97.9&93.7&90.7   &\underline{86.6}         &\underline{84.7}     &\textbf{84.3}     &\underline{81.9}      &\underline{80.4}     &\underline{80.2}     &\underline{79.3}\\
\hline
\textbf{Ours}       &97.8  &\underline{93.8} &\textbf{91.5}   &\textbf{87.5}   &\textbf{85.7}    &\underline{84.2}     &\textbf{82.9}     &\textbf{81.3}     &\textbf{81.5}  &\textbf{80.7}\\
\hline
\end{tabular}}
\caption{\label{main results} Comparison of accuracy (\%) results after learning each task. All models are tested under the same sequences, and relations are equally divided into ten different task sets. The top-performing results are highlighted in bold, and the second-best results are underlined.}
\end{table*}

\section{Experiments}

\subsection{Experimental Settings}
\subsubsection{Datasets}
\textbf{FewRel}~\cite{han2018fewrel} is a few-shot learning relation extraction dataset with 100 relations and 700 instances for each relation. For CRE research, all prior works use 80 relations and divide them into 10 subgroups to replicate 10 distinct tasks.

\textbf{TACRED}~\cite{zhang2017position} is a news network and online documents relation extraction dataset with 42 relations and 106264 samples. Following previous research, we remove \textit{no relation} label and limit the maximum of 320 train samples and 40 test samples for each relation in our experiments. Relations are also divided into 10 distinct portions and are learned by the model continuously.

After $T_k$ is completed, the memory space could include 10 samples for each relation as memory samples.

\subsubsection{Evaluation Metric}
To measure the model effectiveness on all testing sets, we use Accuracy (\%) as the metric. Since the task sequence would affect the midway results of CRE, we construct 5 different task sequences, and the experiment on all open source baselines is under the same task sequence as\cite{cui2021refining,zhao2022consistent,wang2022learning,zhao2023improving} for a fair comparison. Finally, the average results of 5 sequences are taken to compare all models.

\subsubsection{Baselines}
We evaluate our model with the following baselines:
(1) \textbf{EA-EMR}~\cite{wang2019sentence}: uses an explicit embedding alignment model by regularization term through model variation. (2) \textbf{EMAR}~\cite{han2020continual}: retains memory samples and introduces reconsolidation mechanism for continual relation extraction. (3) \textbf{CML}~\cite{wu2021curriculum}: proposes a curriculum-meta learning method, that aims to apart the difficulty of learning different samples. (4) \textbf{RP-CRE}~\cite{cui2021refining}: adds attention module to refine sample embedding with prototypes during the replay learning period. (5) \textbf{CR-ECL}~\cite{hu2022improving}: trains samples with the closest prototypes additionally by margin loss while replaying. (6) \textbf{ACA}~\cite{wang2022learning} increases adversarial class
augmentation mechanism during initial training artificially to enhance the robustness of the system. (7) \textbf{CRL}~\cite{zhao2022consistent} utilizes contrastive learning and knowledge distillation to alleviate catastrophic forgetting. (8) \textbf{CEAR}~\cite{zhao2023improving} combines cross-entropy loss and contrastive loss, uses data augmentation, and focus-loss on memorized samples.

\subsection{Main Results}
\label{Main experiment}
Table~\ref{main results} shows the performances of DP-CRE and all other baselines. Our DP-CRE outperforms previous CRE work, improving 0.7/1.4 accuracy at $T_{10}$ on FewRel and TACRED datasets. The TACRED dataset poses a significant challenge for CRE work due to the class imbalance and the smaller number of training samples available for each relation. Despite these difficulties, our model has managed to make further improvements on this challenging task. This is a testament to the effectiveness of our approach and its ability to handle complex and nuanced relations between entities.
It is worth noting that our work has revealed a more significant enhancement in the later CRE tasks, upon more in-depth analysis. We think as the task rounds increase, the feature space becomes denser and the number of new and prior tasks becomes more imbalanced, DP-CRE can accumulate advantages because of more accurate training in each round. Our technique for controlling changes and balancing tasks can improve the scalability and stability of the model.

\begin{table}[!t]
\centering
\scalebox{0.8}{
\begin{tabular}{l|cc}
\hline 
&\textbf{FewRel}&\textbf{TACRED}\\
\hline
\textbf{Intact Model}   &\textbf{85.1} & \textbf{80.7}\\
\hline
w/o IN &83.7&75.4 \\
\hline
w/o DP &84.4&80.2\\
w/o CA &84.7&79.2\\
w/o BA &84.9&80.1\\
\hline
w/o D-NCM&84.4&79.7\\
\hline
\end{tabular}}
\caption{\label{T10 Ablation} Final $T_{10}$ accuracy(\%) results of ablation experiment. We remove initial learning(IN) at the initial learning step, decoupled contrastive learning(DP), change amount limitation(CA), and multi-task balance(BA) at the replay learning step, and double-NCM prediction(D-NCM) for prediction.}
\vspace{-0.6cm}
\end{table}

\begin{figure}[!h]
\centering
\vspace{-0.3cm}
\subfloat[FewRel Ablation Results]{ \includegraphics[width = 0.30\textwidth]{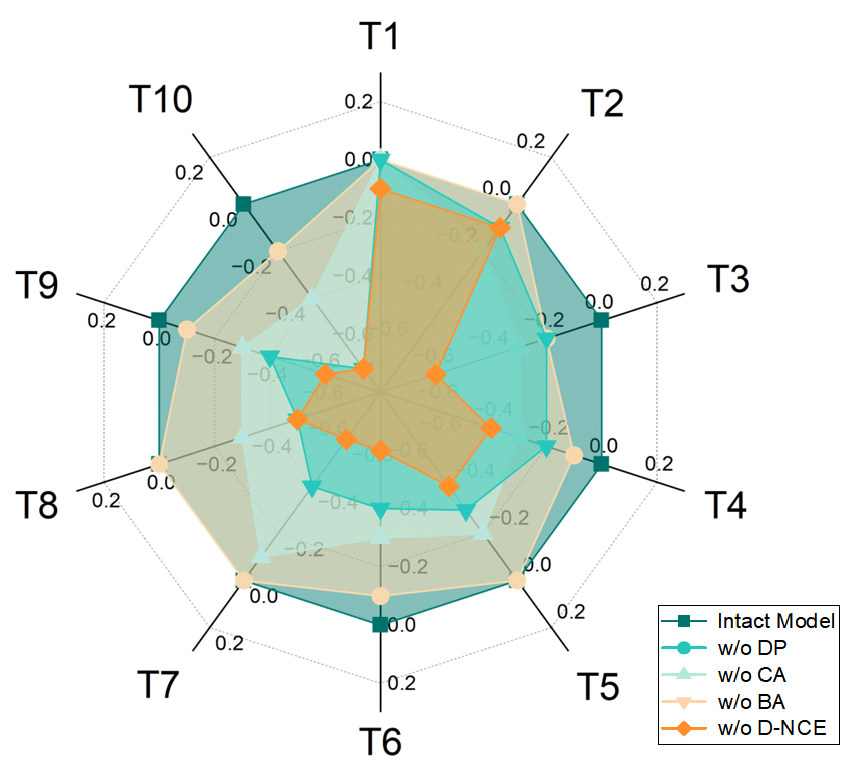}}
\hfill\\\vspace{-0.3cm}
\subfloat[TACRED Ablation Results]
{\includegraphics[width = 0.30\textwidth]{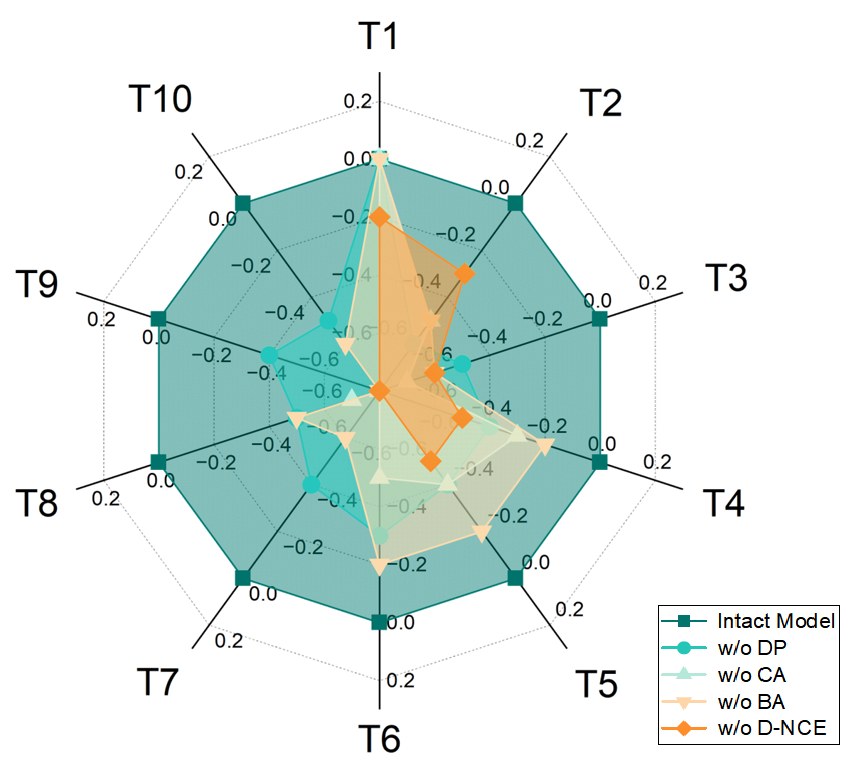}}
\vspace{-0.3cm}
\caption{\label{ablation study} All ablation study results. We calculate $\Delta$ accuracy (\%) between all ablation settings and intact models as table~\ref{T10 Ablation} for each round. }
\vspace{-0.6cm}
\end{figure}

\begin{table}[!h]
\centering
\scalebox{0.7}{\begin{tabular}{l|ccc}
\hline
\multicolumn{4}{c}{\textbf{FewRel}}\\
\hline 
\multicolumn{1}{c}{Memory Size}&5       &10      &15\\
\hline
ACA~\cite{wang2022learning}    &82.8    &84.4    &85.1\\
CRL~\cite{zhao2022consistent}  &80.3    &83.0     &84.0\\
CEAR~\cite{zhao2023improving}  &82.6    &84.0    &84.9\\
\hline
\textbf{Ours}   &\textbf{83.4} &\textbf{85.1}&\textbf{86.1}\\
\hline
\hline
\multicolumn{4}{c}{\textbf{TACRED}}\\
\hline
\multicolumn{1}{c}{Memory Size}&5       &10      &15\\
\hline
ACA~\cite{wang2022learning}    &76.2    &77.5    &78.7\\
CRL~\cite{zhao2022consistent}  &75.0    &78.5    &79.7\\
CEAR~\cite{zhao2023improving}  &76.7    &79.3    &80.4\\
\hline
\textbf{Ours} &\textbf{77.3}  &\textbf{80.7} &  \textbf{81.3}\\
\hline
\end{tabular}}
\caption{\label{memory size results} We compare the final accuracy (\%) after $T_{10}$ training when changing the memory size with several strong models.}
\vspace{-0.6cm}
\end{table}

\begin{table*}[!ht]
\centering
\scalebox{0.85}{\begin{tabular}{lc|ccccccccc}
\hline
\multicolumn{11}{c}{\textbf{FewRel}}\\
\hline 
Model & Task&$T_2$  &$T_3$  &$T_4$  &$T_5$   &$T_6$  &$T_7$  &$T_8$   &$T_9$  &$T_{10}$\\
\hline 
ACA    &old&1.50  &2.50  &2.86  &3.29  &3.85   &4.35  &5.09  &5.09  &5.48\\
&new&1.33&2.03&3.06&3.13&4.69&4.06&5.31&6.34&5.53\\
CEAR          &old&1.41&2.08&2.64&3.11&3.49&4.23&4.70&5.48&6.16\\
&new&1.08&1.80&\textbf{2.16}&\textbf{2.81}&3.66&3.41&4.94&5.59&5.13\\
\hline
\textbf{Ours} &old&\textbf{1.22}&\textbf{1.67}&\textbf{2.29}&\textbf{2.91}&\textbf{3.27}&\textbf{3.73}&\textbf{3.52}&\textbf{4.48}&\textbf{4.70}\\
&new&\textbf{0.96}&\textbf{1.63}&2.26&2.91&\textbf{3.42}&\textbf{3.09}&\textbf{4.78}&\textbf{5.34}&\textbf{4.59}\\
\hline
\hline
\multicolumn{11}{c}{\textbf{TACRED}}\\
\hline
Model & Task&$T_2$  &$T_3$  &$T_4$  &$T_5$   &$T_6$   &$T_7$  &$T_8$ &$T_9$ &$T_{10}$\\
\hline
ACA          &old&1.83  &2.92  &3.15  &3.39  &3.96   &4.43  &4.89  &5.13  &5.47\\
             &new&1.30  &2.33  &3.05	&3.10	&4.75	&4.18  &5.28  &5.85 &5.85\\
CEAR         &old&1.33 &2.15&2.71&3.21&3.58&4.30&4.58 &5.43 &5.98\\
&new&\textbf{0.90}  &2.00  &2.18  &\textbf{2.78}  &\textbf{3.73} &3.25 &5.13  &5.05  &5.45\\
\hline
\textbf{Ours} &old&\textbf{1.08} &\textbf{1.99}  &\textbf{2.54} &\textbf{3.04}&\textbf{3.42}  &\textbf{3.82} &\textbf{4.35}   &\textbf{4.61}  &\textbf{4.78}\\
&new&1.07 &\textbf{1.80}   &\textbf{2.08} &2.95	 &\textbf{3.73}	&\textbf{3.13}	&\textbf{4.90}	&\textbf{5.00}	&\textbf{5.10}\\
\hline
\end{tabular}}
\caption{\label{Delta F1} The average $\Delta F1$(\%) between old and new tasks. In each row of a model, the top line represents $\Delta F1$ of the old tasks and the bottom line represents the new tasks. }
\vspace{-0.6cm}
\end{table*}

\subsection{Ablation Study}
This part aims to test the effectiveness of individual modules of the DP-CRE framework. The results are presented in figure~\ref{ablation study}. In "w/o IN", we removed the initial learning step. In "w/o DP", we replaced decoupled contrastive learning ($L_{DPCon}$) with supervised contrastive learning ($L_{SupCon}$). In "w/o CA" and "w/o BA", we removed the change amount limitation that restricts memory sample embedding, and the module used to calculate the balance coefficient between new and old tasks by setting the balance coefficients $\gamma = 0.5$. In the "w/o D-NCE" experiment, we used average prototypes to predict the test samples.
Our research demonstrates the efficiency and necessity of our model by showing how each component contributes to its improvement. Additionally, from table~\ref{T10 Ablation}, we observed that the CA-Limit module displayed more improvements on the TACRED dataset. We think it is primarily because TACRED consists of a larger number of conflicting relation types, making CA-Limit more significant in handling frequent embedding changes.

\subsection{Influence of Memory Size}
In this experiment, we change the memory size to verify the universality of model lifting. All other settings are the same as experiment~\ref{Main experiment} except memory space size is set to 5 or 15 for each relation. The final accuracy after $T_{10}$ is presented in table~\ref{memory size results}. It is observed that the accuracy of the experimental results increased as the memory size increased due to the additional memory samples providing more information from old tasks during replay learning. We compare our model with several strong baselines and find that our method of fully utilizing memory samples results in the best outcome of the CRE task. The improvement is more significant when more memory samples are saved, due to our replay strategy focusing on the change amount for each memory sample individually.

\subsection{Task Balance Experiment}
To validate the effectiveness of our balancing strategy, we conduct a task balance experiment and compare DP-CRE with two strong baselines. Table~\ref{Delta F1} shows the average predictive accuracy of new and old tasks calculated separately in the same round.
There are two main reasons for the decrease in the F1 value. The first reason is the confusion caused by additional new relation labels. The second reason is the catastrophic forgetting caused by CRE. For example, from the perspective of a relation, the F1 value of the \textit{P137operator} relation decreases by the same amount as the regular all-data-available RE task results. However, \textit{P937work location} experiences a sudden drop that only appears in the CRE experiment, which means catastrophic forgetting. To assess the impact of the CRE task, we use the F1 difference $\Delta F1$ of the CRE model and the regular RE model.
Our approach effectively improves the performance of the old and new tasks and achieves the best performance on all old tasks. Perhaps in some rounds, DP-CRE does not achieve optimal results on the new task experiment. We think it is to prevent any over-bias towards either side in case of conflicts, thereby ensuring a balanced model.

\vspace{-0.3cm}
\section{Conclusion}
This paper proposes a DP-CRE framework to balance prior information preservation and new knowledge acquisition. During the training process, we monitor the changes in model embedding and control the model with a change amount to maintain the structural information of memory samples. The experimental results demonstrate that DP-CRE can significantly enhance the performance of state-of-the-art CRE models.
Our model also has two limitations. Firstly, compared with the invariant embedding model, this model has advantages when the embedding space is fuller, which means we can conduct deep research on embedding changes. Secondly, although we view CRE as multi-task learning, the processing in this paper is a general continual learning strategy. 
We leave it as further work that integrates with the specifics of RE tasks.

\section{Acknowledgements}
This work was supported by the National Key Research and Development Plan of China under Grant No. 2022YFF0712200 and 2022YFF0711900, the Natural Science Foundation of China under Grant No. T2322027, the Postdoctoral Fellowship Program of CPSF under Grant No. GZC20232736, the China Postdoctoral Science Foundation Funded Project under Grant No.2023M743565, Information Science Database in National Basic Science Data Center under Grant No.NBSDC-DB-25, Youth Innovation Promotion Association CAS.

\nocite{*}
\section{Bibliographical References}\label{sec:reference}

\bibliographystyle{lrec_natbib}
\bibliography{lrec-coling2024-example}

\bibliographystylelanguageresource{lrec_natbib}
\clearpage

\begin{table}[!h]
\centering
\scalebox{0.8}{\begin{tabular}{l|cc}
\hline 
&\textbf{FewRel}&\textbf{TACRED}\\
\hline
Baseline&84.0&78.7\\
\hline
+ DP &84.4&79.3\\
+ CA &84.3&80.2\\
+ D-NCM &84.4&80.0\\
\hline
\textbf{Intact Model}   &\textbf{85.1} & \textbf{80.7}\\
\hline
\end{tabular}}
\caption{\label{Single Module} Final $T_{10}$ accuracy(\%) results. We compare a baseline model with each module added individually, including decoupled contrastive learning(DP), change amount limitation(CA), and double-NCM prediction(D-NCM) for prediction. }
\vspace{-0.6cm}
\end{table}


\appendix
\section{Additional Ablation}
To more clearly demonstrate the contribution of each module in the ablation study and avoid doubt about better performance in the baseline model, we conduct additional ablation experiments individually for each module in table~\ref{Single Module}. The baseline model includes no additional modules. Since Multi-task balance(BA) is the balance of two modules, it cannot be added separately. The experimental results provide more evidence of the effectiveness of the DP-CRE modules.

\begin{figure}
\centering
\subfloat[TACRED]{ \includegraphics[width = 0.38\textwidth]{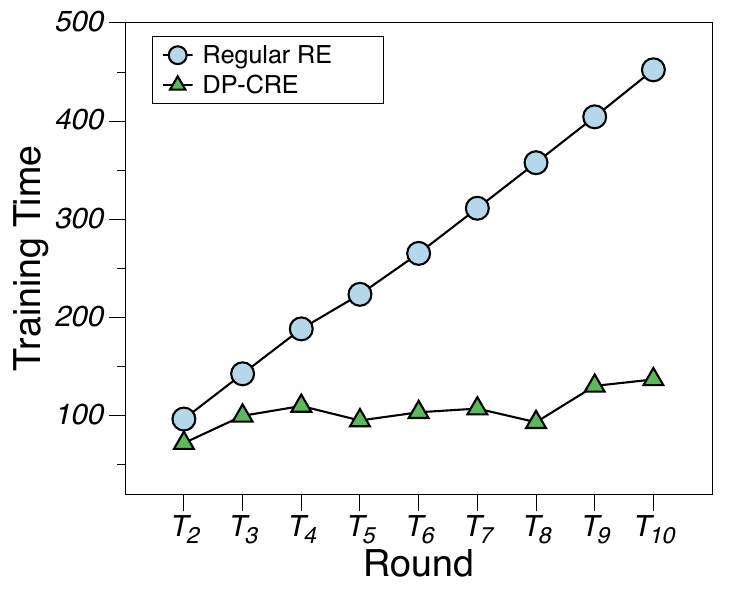}}
\\\vspace{-0.3cm}
\subfloat[FewRel]
{\includegraphics[width = 0.38\textwidth]{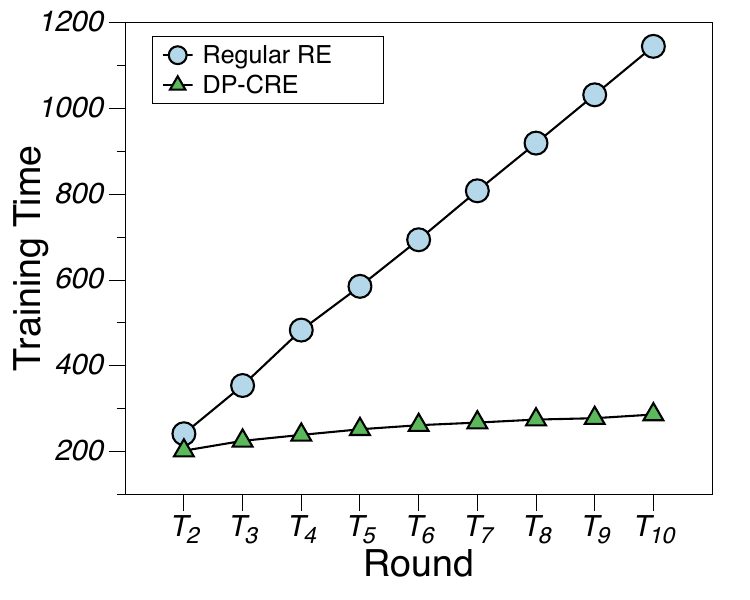}}
\caption{\label{Time} Total training time(s) of DP-CRE and Regular RE. Regular RE is trained using the entire data with the same model architecture.}
\label{fig. change amount}
\end{figure}



\begin{figure}
\centering
\subfloat[TACRED]{ \includegraphics[width = 0.38\textwidth]{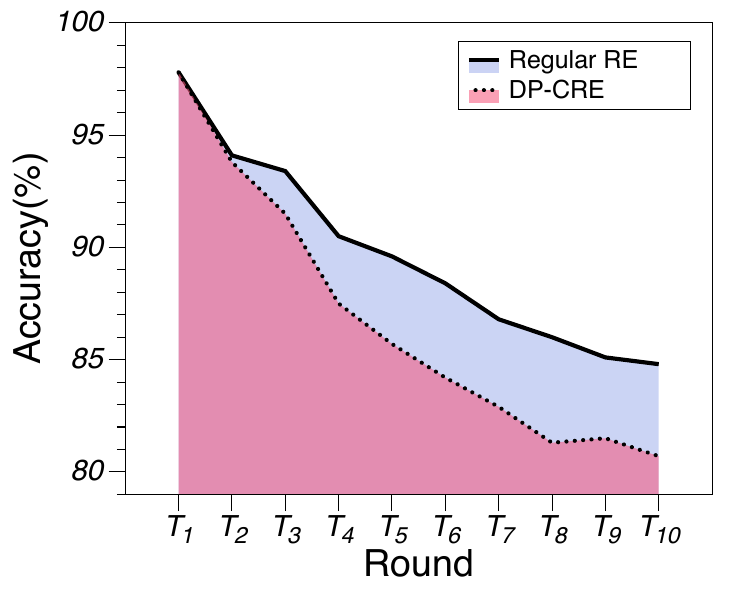}}
\\\vspace{-0.3cm}
\subfloat[FewRel]
{\includegraphics[width = 0.38\textwidth]{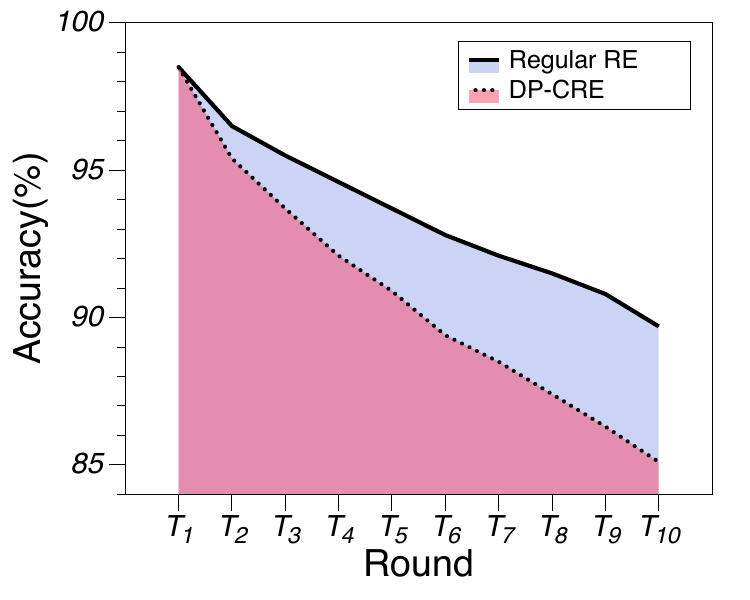}}
\caption{\label{Regular Accuracy} Accuracy (\%) when training the same model architecture with the entire data.}
\label{fig. change amount}
\end{figure}

\section{Training Time}
We additionally conducted training time experiments to verify the advantages of continual learning compared with the regular approach. We find that DP-CRE could reduce the training time and lower the cost of model training significantly, for example, the training time could reduce from 1145.33s/452.26s to 286.31s/137.01s on FewRel and TACRED datasets at $T_{10}$, with a minor reduction in accuracy from 89.7/84.8 to 85.1/80.1 compared to regular RE training. Figure~\ref{Time} shows the whole experiment results. In figure~\ref{Regular Accuracy}, we attach the accuracy of Regular RE in the table for a clear illustration.

\section{Memory Size}
\begin{figure*}
\centering 
\subfloat[FewRel: ACA~\cite{wang2022learning}]
{\includegraphics[width = 0.5\textwidth]{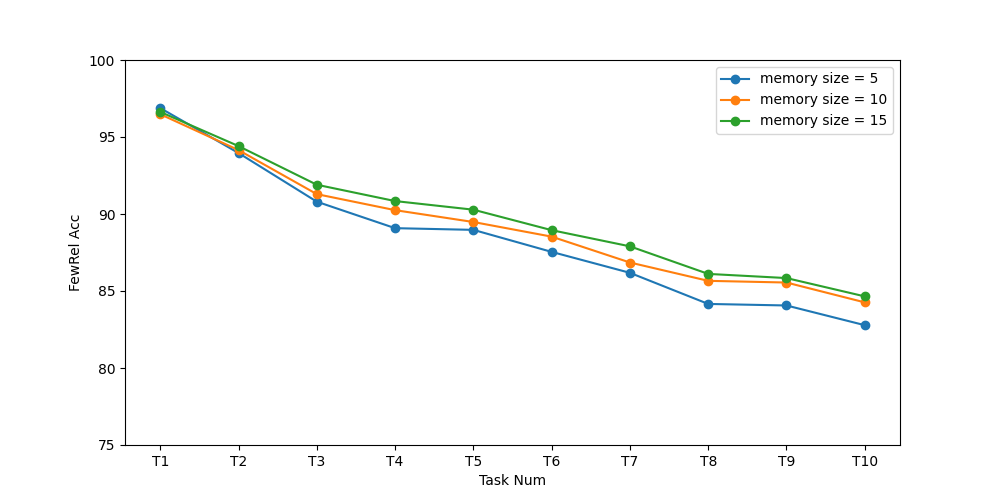}}
\hfill
\subfloat[TACRED: ACA~\cite{wang2022learning}]
{\includegraphics[width = 0.5\textwidth]{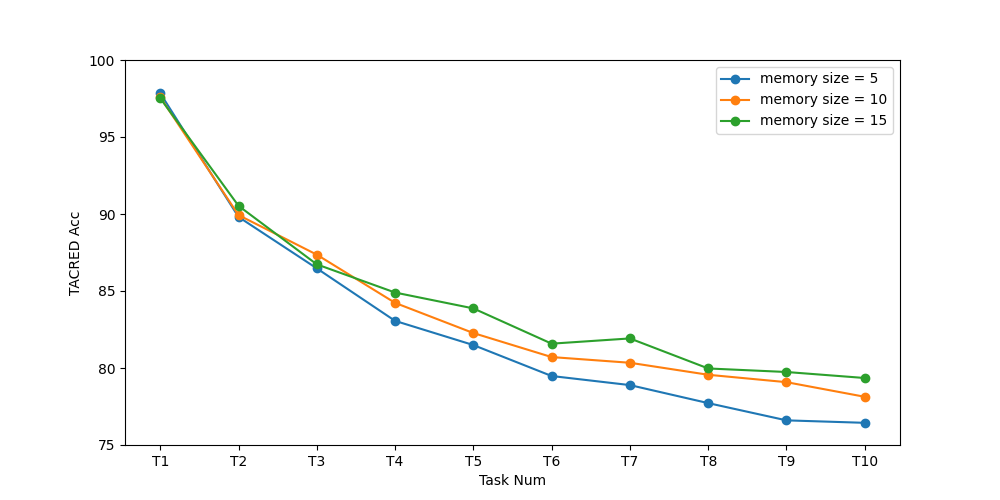}}
\hfill
\subfloat[FewRel: CRL~\cite{zhao2022consistent}]
{\includegraphics[width = 0.5\textwidth]{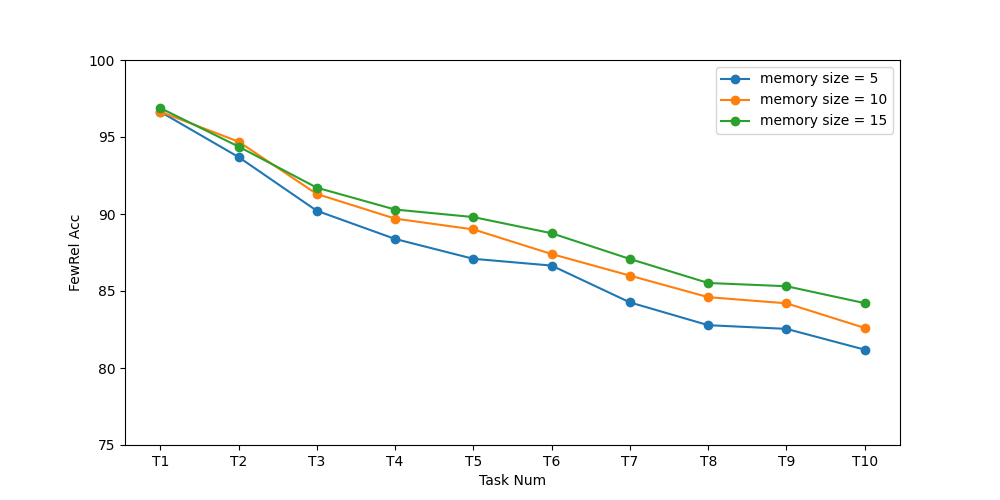}}
\hfill
\subfloat[TACRED: CRL~\cite{zhao2022consistent}]
{\includegraphics[width = 0.5\textwidth]{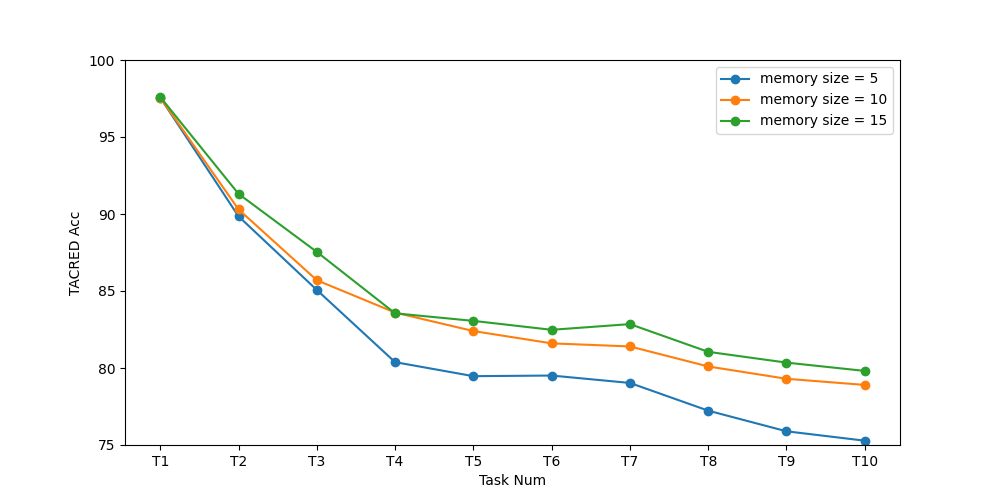}}
\hfill
\subfloat[FewRel: CEAR~\cite{zhao2023improving}]
{\includegraphics[width = 0.5\textwidth]{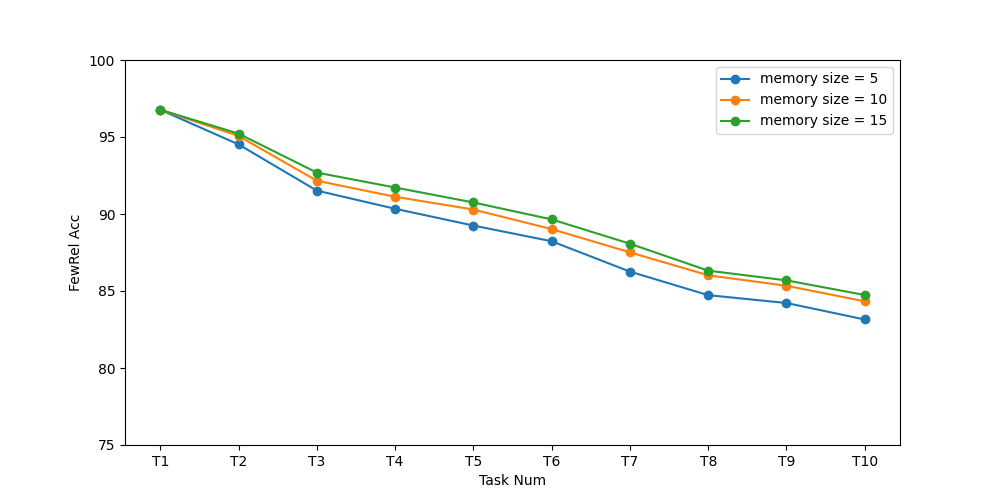}}
\hfill
\subfloat[TACRED: CEAR~\cite{zhao2023improving}]
{\includegraphics[width = 0.5\textwidth]{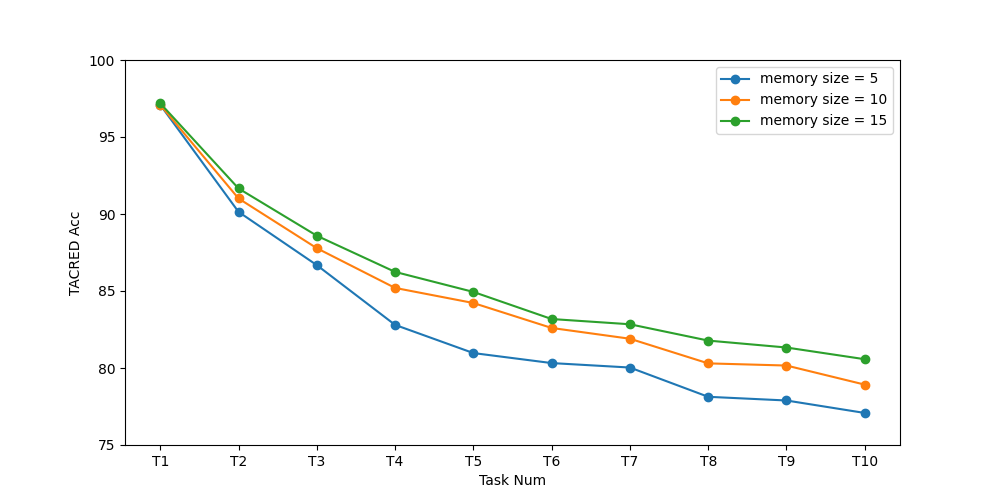}}
\hfill
\subfloat[FewRel: DP-CRE]
{\includegraphics[width = 0.5\textwidth]{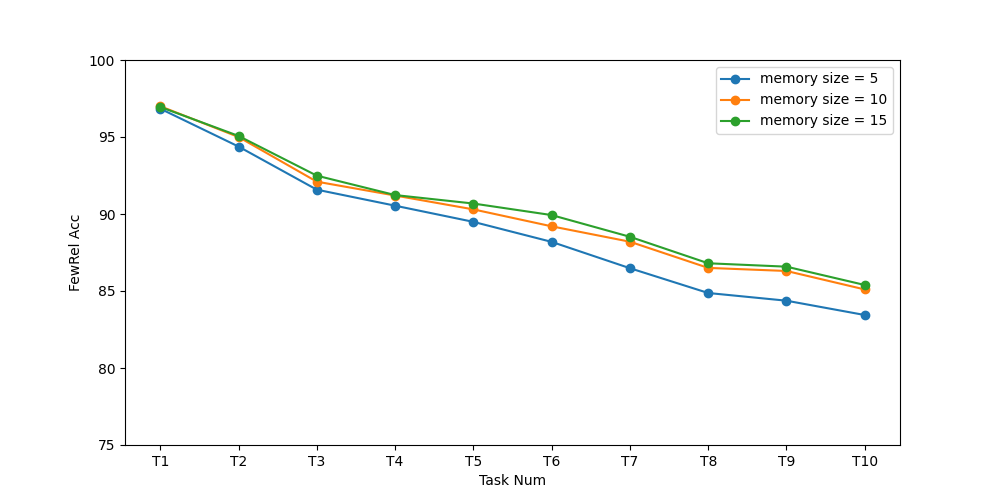}}
\hfill
\subfloat[TACRED: DP-CRE]
{\includegraphics[width = 0.5\textwidth]{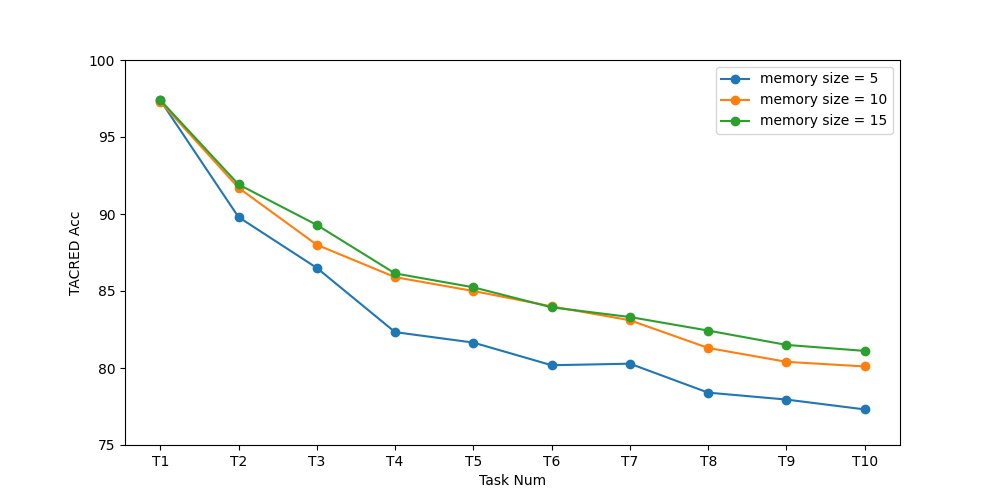}}
\hfill
\caption{The complete memory experiment result. We experimented with several recent models. Each graph includes ten tasks with accuracy(\%) for the same model when changing memory space size. }
\label{fig. complete memory}
\end{figure*}

This part is the complete result of the influence of memory size. All memory size experiment results are shown in figure~\ref{fig. complete memory}. Every graph includes 3 lines of memory size = 5, 10, 15 of one model in one dataset. It is evident that the accuracy rises when increasing memory size and declines when adding training rounds across all models and datasets.

\end{document}